\documentclass[sigconf]{acmart}
\usepackage[noend]{algpseudocode}
\usepackage{algorithm,algorithmicx}

\newcommand{\R}{\mathbb{R}}

\copyrightyear{2022}
\acmYear{2022}
\setcopyright{acmcopyright}
\acmConference[SBST'22]{The 15th Search-Based Software Testing Workshop}{May 9, 2022}{Pittsburgh, PA, USA}
\acmBooktitle{The 15th Search-Based Software Testing Workshop (SBST'22), May 9, 2022, Pittsburgh, PA, USA}
\acmPrice{15.00}
\acmDOI{10.1145/3526072.3527522}
\acmISBN{978-1-4503-9318-8/22/05}

\author{Jarkko Peltomäki}
\affiliation{%
  \institution{Åbo Akademi University}
  \city{Turku}
  \country{Finland}}
\email{jarkko.peltomaki@abo.fi}
\author{Frankie Spencer}
\affiliation{%
  \institution{Åbo Akademi University}
  \city{Turku}
  \country{Finland}}
\email{frankie.spencer@abo.fi}
\author{Ivan Porres}
\affiliation{%
  \institution{Åbo Akademi University}
  \city{Turku}
  \country{Finland}}
\email{ivan.porres@abo.fi}

\title{Wasserstein Generative Adversarial Networks for Online Test Generation for Cyber Physical Systems}

\begin{abstract}
We propose a novel online test generation algorithm WOGAN based on Wasserstein Generative Adversarial Networks. WOGAN
is a general-purpose black-box test generator applicable to any system under test having a fitness function for
determining failing tests. As a proof of concept, we evaluate WOGAN by generating roads such that a lane assistance system of a
car fails to stay on the designated lane. We find that our algorithm has a competitive performance respect to previously
published algorithms.
\end{abstract}

\begin{document}

\maketitle

\section{Introduction}
% Ivan

% Motivation
%% Safety of cyber physical Systems
Safety validation is the process of establishing the correctness and safety of a system operating in an environment \cite{DBLP:journals/jair/CorsoMKLK21}. Validation is necessary to ensure that a system works as expected before it is taken into production. This is especially important with safety properties of cyber-physical systems (CPS), where a fault can lead to severe damage to property or even death.

System validation is a challenging and open problem. While there is a large body of knowledge on the subject, the size, complexity, and requirements for software-intensive systems have grown  over time. There is a need for cost and time efficient approaches that support system-level verification and validation of complex systems including program code, machine learning models, and hardware components. 

% The problem to solve

In this article, we present a novel test generation algorithm for the CPS testing competition organized in 2022 within the 14th International Search-Based Software Testing Workshop (SBST 2022).

The system under test (SUT) is the lane assist feature of a car. This is a feature that aims to  maintain a car safely within the boundaries of its driving lane without human assistance. If working properly, it can help to avoid an accident due to a distracted or tired driver. On the other hand, a faulty implementation may fail to avoid an accident or even actively cause one.

The main safety requirement for the lane assist feature is described as an upper bound $b$ on the percentage of the body of the car that is out of the boundaries of its lane ($ \mathsf{BOLP}$) at any given moment. We can formalize this in signal temporal logic~\cite{DBLP:conf/rv/Donze13} as $\square \mathsf{BOLP} \leq b$. The goal of the test generator is to falsify this requirement. 

\begin{figure}
  \begin{minipage}{0.49\columnwidth}
    \includegraphics[width=\linewidth]{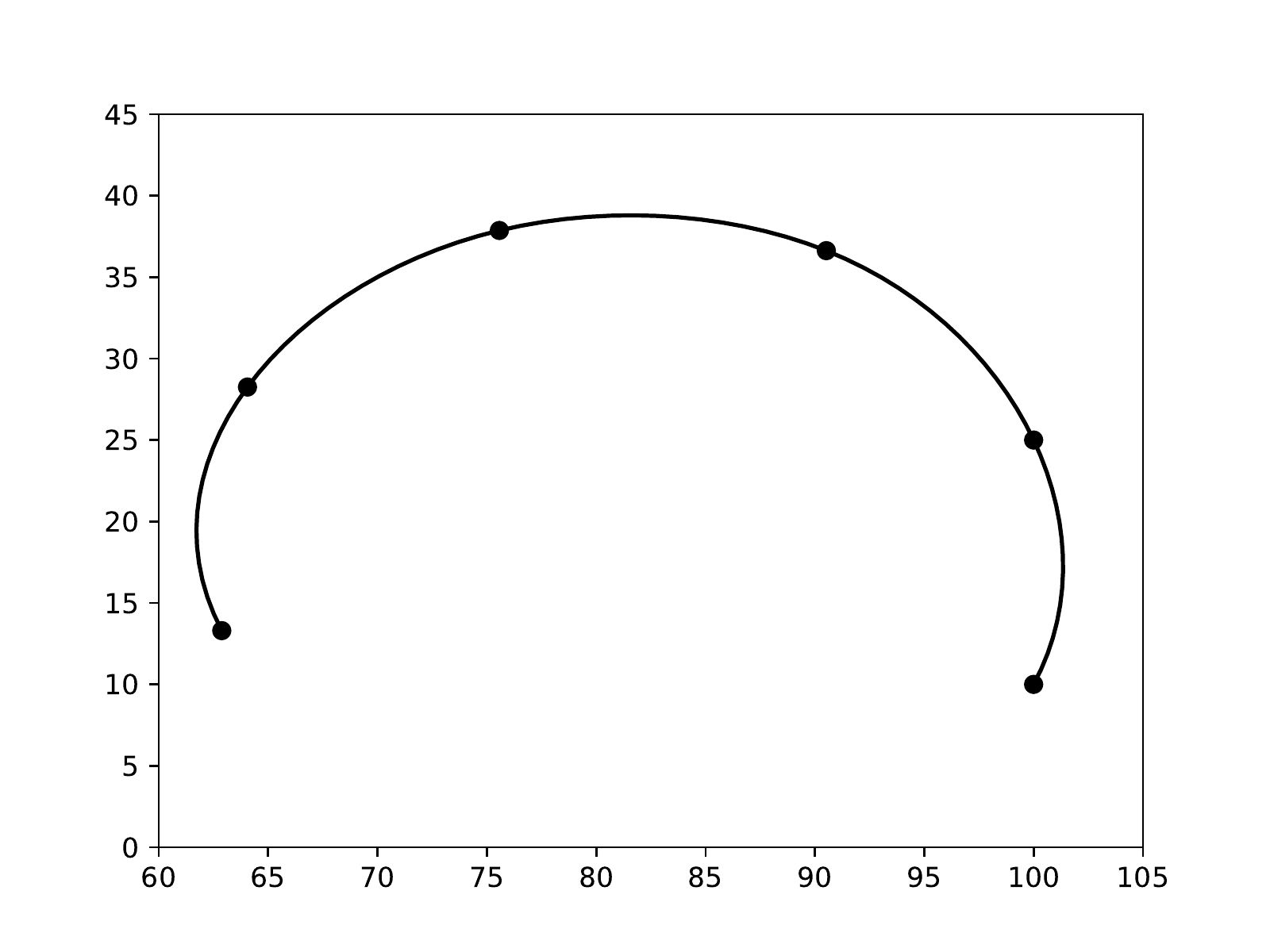}
  \end{minipage}%
  \begin{minipage}{0.49\columnwidth}
    \includegraphics[width=\linewidth]{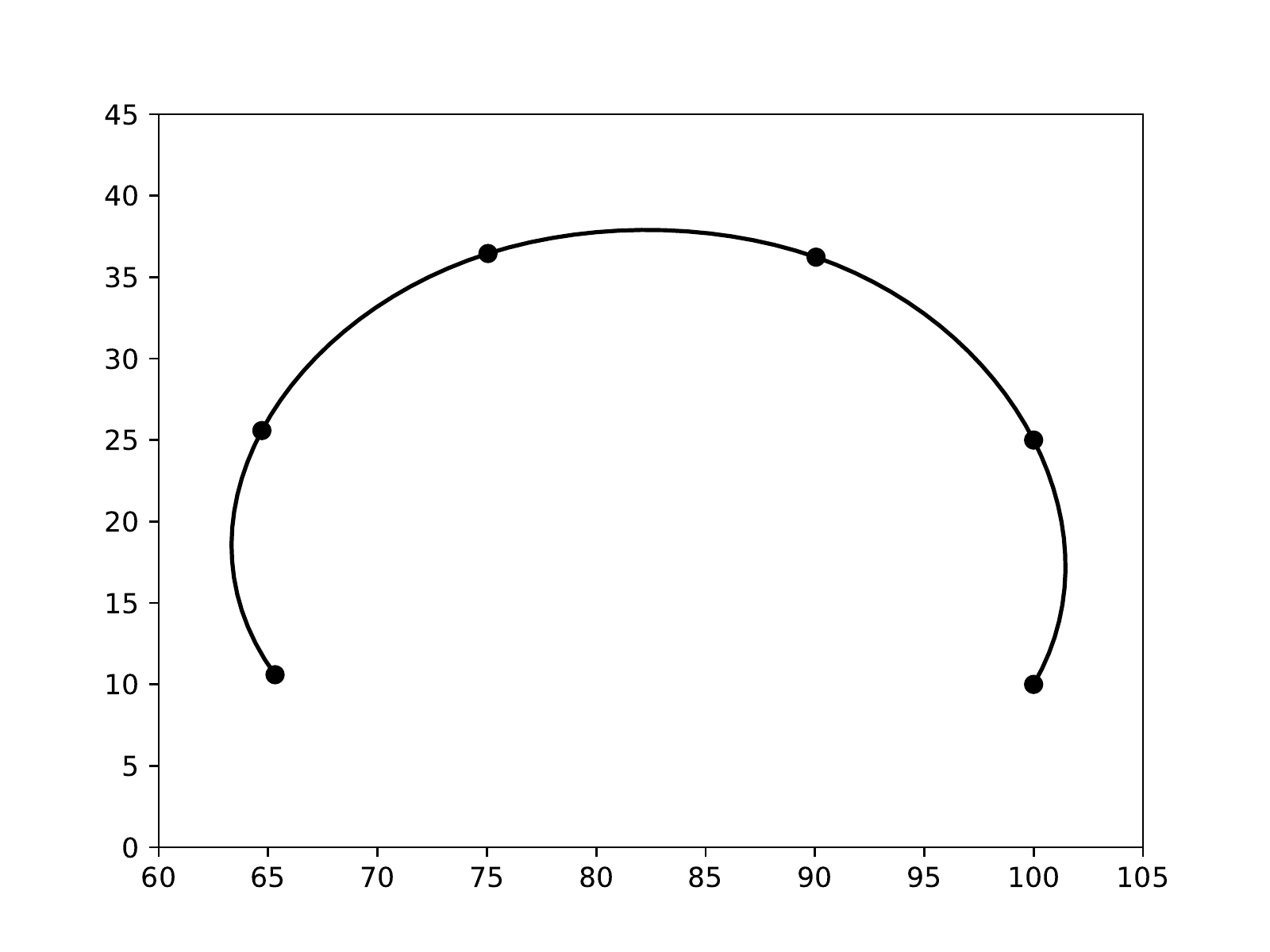}
  \end{minipage}%
  \caption{\textmd{Two visually similar tests having fitnesses $0.26$ (left) and $0.99$ (right).}}\label{fig:roads}
\end{figure}

We do not have access to any kind of specification, design document, or source code of the SUT. Instead, we are provided an implementation that can be executed in a simulated environment and a specification of the inputs and outputs of this simulator. The input of the simulator is a driving scenario defined as a road with the car as the only traffic. The output of a test $t$ is the maximum $\mathsf{BOLP}$ observed during the simulation, which we denote by $f(t)$. Based on this premise, our approach is in essence a system-level, black-box online test generator. It produces roads as inputs, observes the system outputs, and uses these observation to decide what should be the next test to execute. The goal is to find a failing test, i.e., a test $t$ whose fitness $f(t)$ satisfies $f(t) > b$. In order to appreciate how difficult the task is, we included in \autoref{fig:roads} two roads defined by $6$ points which are visually similar but have very different fitnesses. This also shows that humans have hard time estimating the lane assist performance: a human driver would perform equally well on both roads.

There are two main quality criteria for our test generator. First, our generator should aim to generate as many failing tests as possible within a given time budget. Executing a simulation is a time consuming process and the testing process is limited by a budget given as a time limit.  The second goal is to generate failing tests that are as diverse as possible. The hypothesis is that a set of diverse failing tests can improve the process of determining the root cause of the observed errors.  These are two conflicting criteria since a test generator may generate many failing tests by exploiting previously found positives instead of exploring the input space for more diverse tests.

% Main contribution and novelty
The test generation algorithm presented in this article is based on a novel approach to the falsification of CPS. While most existing approaches use a known metaheuristic such as a genetic algorithm to search for tests with high fitness, our algorithm uses a Wasserstein generative adversarial network (WGAN) \cite{wgan} as a generator for such tests. Our algorithm starts tabula rasa and interacts with the system to learn how to generate high-fitness tests. The algorithm has no domain knowledge about the SUT except for the representation of the input space to facilitate random search and learning by a neural network. Minimal usage of domain knowledge makes our algorithm general-purpose, i.e., it is easily adjusted for other SUTs. Our experimental results indicate that our algorithm has competitive performance when compared with the previously published algorithms \cite{DBLP:conf/sbst/PanichellaGZR21}.

We remark that our algorithm shares ideas with the algorithm of the paper \cite{ogan} which is inspired by GANs. The
algorithm of \cite{ogan} does not train a GAN in the sense of the original GAN paper \cite{goodfellow2014generative} as
it trains a generator neural network whose outputs achieve a large value when fed through a discriminator neural
network. While this approach can achieve good results \cite{ogan}, it is a viable strategy for the generator to always
generate a single good test. In such a case, the resulting test generator would not fulfill the second quality
criterion. In order to work around this potential problem, our algorithm trains a proper GAN. In fact, we train a WGAN,
and WGANs are known to be able to produce varied samples from their target distributions \cite{wgan}.

% Mini-TOC
We present our test generation algorithm in detail in \autoref{sec:algorithm} while its performance in test generation tasks similar to the SBST 2021 CPS testing competition is presented in \autoref{sec:results}. Finally, we present our conclusions and a discussion of future work in \autoref{sec:conclusions}.

\section{A Novel Algorithm}\label{sec:algorithm}
% Jarkko
\subsection{Feature Representation}\label{ssec:feature_representation}
The input to the simulator is a sequence of points in the plane which the provided simulator interface interpolates to
a road. See \autoref{fig:roads} for roads defined by $6$ points. Not all sequences are valid: intersecting roads and
roads with steep turns are disallowed. Moreover, a road must fit in a map of $200 \times 200$ units. An efficient
black-box validator for candidate roads is provided by the SBST 2021 CPS competition
\cite{DBLP:conf/sbst/PanichellaGZR21}.

Generating valid tests by randomly choosing sequences of plane points is difficult, so we opted to use the feature
representation described in \cite{frenetic}. For us, a test is a vector in $\R^d$ whose components are curvature values
in the range $[-0.07, 0.07]$. Given a test, we fix the initial point to be the bottom midpoint of the map and the
initial direction to be directly up. Then we numerically integrate the curvature values with a fixed step length $15$ to
obtain $d$ more points. Fixing the initial point and direction is justified by noting that the test fitness (maximum
$\mathsf{BOLP}$) should be translation and rotation invariant.

This feature representation makes random search of valid roads more feasible. In our experiments, we set $d = 5$ and we
estimated that then the probability of a test being valid equals $0.48$. On the other hand, we estimated that a
sequence of $6$ plane points with components chosen uniformly randomly in $[0, 200]$ corresponds to a valid road with
probability $0.0036$. For both estimates, we generated $5000$ random roads.

\subsection{Overview of Models and Their Training}
Let $\mathcal{T}$ be the set of failing tests in the test space $[-0.07, 0.07]^d$. Our aim is to learn online a
mapping $G$ from a latent space $[-1, 1]^{d'}$ to the test space such that uniform sampling of the latent space yields
samples of $\mathcal{T}$ via the map $G$. To achieve this, we train this generator $G$ as a WGAN \cite{wgan}. In this
setting, $G$ is a neural network and it is trained to minimize the Wasserstein distance between the distribution of $G$
and the real data distribution $\mathcal{T}$. This is accomplished with the help of a critic $C$ (also a neural
network) which, informally speaking, is trained to distinguish between tests in $\mathcal{T}$ and outside of it.

Training the WGAN traditionally requires having a training data sample from $\mathcal{T}$ in advance. Since we do not
have such data, an alternative approach is needed. We propose to begin from a set $\mathcal{R}$ of random valid tests
and update it using an analyzer $A$ (also trained online). We execute the tests of $\mathcal{R}$ on the SUT to know
$f(t)$ for all $t \in \mathcal{R}$. We use this data to train $A$ (a neural network) to approximate the mapping $f$. We
may now use the generator $G$ to produce valid tests and estimate their fitness by $A$ without consulting the SUT
(recall that a validator is available). When we have found a valid test which is estimated by $A$ to have high fitness,
we execute it on the SUT and add the test and its outcome to $\mathcal{R}$.

Training the WGAN directly on $\mathcal{R}$ treats all tests equally which is counterproductive as $\mathcal{R}$ can
contain many low-fitness tests. We propose to create a training batch from $\mathcal{R}$ which is biased towards
high-fitness tests and train the WGAN on it. Ideally $G$ becomes more able to generate high-fitness tests and $A$ gets
more accurate as more data is available. The analyzer $A$ does not need to be perfect; it just needs to drive the
training towards high-fitness tests.

It has been observed that WGANs, like GANs, can produce novel variations of their  training data \cite{wgan} so,
whenever the fitness function $f$ is well-behaved on the test space (e.g., it is locally continuous) and the random
search is representative enough, the generator should eventually produce high-fitness tests, some of which belong to
$\mathcal{T}$. Failing this could be taken as an indication that faults do not exist.

Our experiments confirm that \autoref{alg} can expose many faults of the SUT of the SBST 2021 CPS testing competition
showing empirically the validity of the above ideas. This achieves the first quality criterion of a test generator.
WGANs have been shown to avoid the so-called GAN mode collapse in which a generator produces the same output over and
over \cite{wgan}. We observe this in our experiments as we typically find clusters of failing tests that are largely
different from each other. This achieves the second quality criterion.

\begin{algorithm}
  \caption{WOGAN online test generator}\label{alg}
  \begin{algorithmic}[1]
    \Require{Execution time budget time\_budget, number of initial random tests $N$, multiplier target\_reducer.}
    \Statex

    \State $A \gets$ \Call{initialize\_analyzer}{ }
    \State $G, C \gets$ \Call{initialize\_wgan}{ }
    \State $\alpha \gets$ \Call{initialize\_batch\_parameter}{ }
    \State $T, F \gets$ \Call{sample\_and\_execute\_random\_tests}{$N$}
    \While{clock $<$ time\_budget}
      \State \Call{train\_analyzer}{$A; T, F$}
      \State $X \gets$ \Call{sample\_batch}{$T, F; \alpha$}
      \State \Call{train\_wgan}{$G, C; X$}

      \State target $\gets 1$
      \Repeat
        \State test $\gets$ \Call{generate}{$G$}
        \If{$\lnot$ \Call{valid}{test}}
          \State continue
        \EndIf
        \State target $\gets$ target $\cdot$ target\_reducer
      \Until{\Call{predict}{$A;$ test} $\geq$ target}

      \State test\_fitness $\gets$ \Call{execute}{test}
      \State $T \gets T \, \cup \, \{$test$\}$
      \State $F \gets F \, \cup \, \{$test\_fitness$\}$
      \State $\alpha \gets $ \Call{update\_batch\_parameter}{clock, $\alpha$}
    \EndWhile
\end{algorithmic}
\end{algorithm}

\subsection{The Algorithm}
Let us explain the pseudocode of our WOGAN algorithm \autoref{alg}. Its main structure is similar to the OGAN algorithm
of \cite{ogan}. Initially a population of $N$ random tests is created and executed on the SUT. On each round of the
outer loop, the analyzer $A$ is trained on the current tests $T$ and their fitnesses $F$, and the WGAN consisting of
the generator $G$ and the critic $C$ is trained on a biased batch of samples from $T$ (see the next paragraph). Then
candidate tests are produced by $G$ in the inner loop until a valid test whose fitness $A$ estimates to be high enough
has been produced. The acceptance threshold is lowered on each iteration in order to find suitable candidates
reasonably fast. The accepted test is added to the test suite along with its fitness. Finally a parameter $\alpha$
controlling the biased batching is updated. The algorithm terminates when the time budget is exhausted.

Let us then describe how the biased batches are formed. Partition the interval $[0, 1]$ (the range of $f$) into $B$
contiguous bins of length $1/B$. Put each test in $T$ to a bin according to its fitness. We weight the bins and sample
$M$ bins ($M$ is the batch size) according to this weighting. The biased batch is formed by choosing a test uniformly
randomly (without replacement) from each sampled bin. The weights are chosen according to a shifted sigmoid function
$1/(1 + \exp(-(x+\alpha)))$. The parameter $\alpha$ increases linearly from $0$ to $3$ during the execution. Thus
initially low-fitness tests have a chance of being selected, encouraging exploration, whereas later only high-fitness
tests are likely to be sampled.

\section{Experimental Results}\label{sec:results}
In this section we compare our WOGAN algorithm to two other algorithms: random search (to establish a baseline)
and the Frenetic algorithm \cite{frenetic}, a genetic algorithm whose performance was deemed to be among the best of
the SBST 2021 CPS testing competition entries \cite{DBLP:conf/sbst/PanichellaGZR21}.

\subsection{Experiment Setup}
The goal of the test generation is to generate a test suite with as many failing tests as possible within a time
budget of $2$ hours. A test is considered failed if its fitness (maximum $\mathsf{BOLP}$) exceeds $0.95$. No speed
limit is imposed on the simulator AI. We report the results of $20$ repeated experiments for each algorithm.

We use the version $0.24.0.1$ of the BeamNG.tech simulator as in the current 2022 iteration of the competition. The
lane assist AI used is the one provided by BeamNG.tech. The data was collected on a desktop PC running Windows 10
Education with Intel i9-10900X CPU, Nvidia GeForce RTX 3080 GPU, and 64 GB of RAM. \\

\noindent
\textbf{WOGAN.} We produce roads with $6$ points, i.e., we set $d = 5$. This is an arbitrary decision driven by the
fact that it is challenging to train neural networks with limited training data of high dimensionality. We set
$d' = 10$, so the latent space has dimension $10$. In \autoref{alg}, we set $N = 60$ (approximately $20 \%$ of the
roads that can be tested in the given time budget) and target\_reducer $= 0.95$. We use $B = 10$ ($10$ bins) and $M =
32$ (batch size $32$) in the WOGAN batch sampling. Our random search produces tests $(c_1, \ldots, c_5)$, as described
in \autoref{ssec:feature_representation}, such that $c_{i+1} \in c_i \pm [-0.05, 0.05]$ for all $i \geq 1$. Frenetic
uses the same procedure.

We train the WGAN using gradient penalty as in \cite{wgan_gp}. The critic and generator both use a dense neural network
of two hidden layers with $128$ neurons and ReLU activations. The analyzer has two hidden layers with $32$ neurons and
ReLU activations. The WGAN is trained with the default settings of \cite{wgan_gp} and learning rate $0.00005$ as
suggested in \cite{wgan}. The analyzer is trained with the Adam optimizer with learning rate $0.001$ and beta
parameters $0$ and $0.9$. These settings yielded good performance, but we did not attempt to determine the best
parameters empirically. \\

\noindent
\textbf{Frenetic.} Frenetic \cite{frenetic} is a genetic algorithm that uses, in our opinion, domain-specific mutators
to generate new solutions. A remarkable feature in Frenetic is the use of two sets of mutators depending on if an
individual represents a passed or a failed test. Mutators for passed tests are used to explore the solution space while
the mutators for failed tests exploit the failure to generate new high-fitness roads.  The mutators for failed test
mutators are: reversing the cartesian representation of a road, reversing its curvature values, splitting the road in
two halves and swapping them, and mirroring the road. These mutators create new roads that are diverse by construction
under many similarity measures and may lead to tests with high-fitness.

We updated the Frenetic algorithm to use the latest version of the simulator. We also set the number of roads for the
initial random search to $60$ which is the same as in WOGAN. Due to the changes, we consider that the version of
Frenetic used in this article is not equivalent to the one presented in \cite{frenetic} and its results are not
comparable.

Notice that Frenetic produces roads defined by variable number of points. Its random search builds roads with
$20 \pm 5$ points and the genetic algorithm rules can change this number. As stated above, WOGAN considers only roads
defined by $6$ points which is a much lower number. We opted not to change Frenetic in order to keep our modified
Frenetic as close as possible to the original. \\

\noindent
\textbf{Random.}
This algorithm simply generates uniformly randomly tests $(c_1, \ldots, c_5)$ as described in
\autoref{ssec:feature_representation} (that is, we generate roads of $6$ points).

\subsection{Results}

\begin{figure}
  \includegraphics[width=\columnwidth]{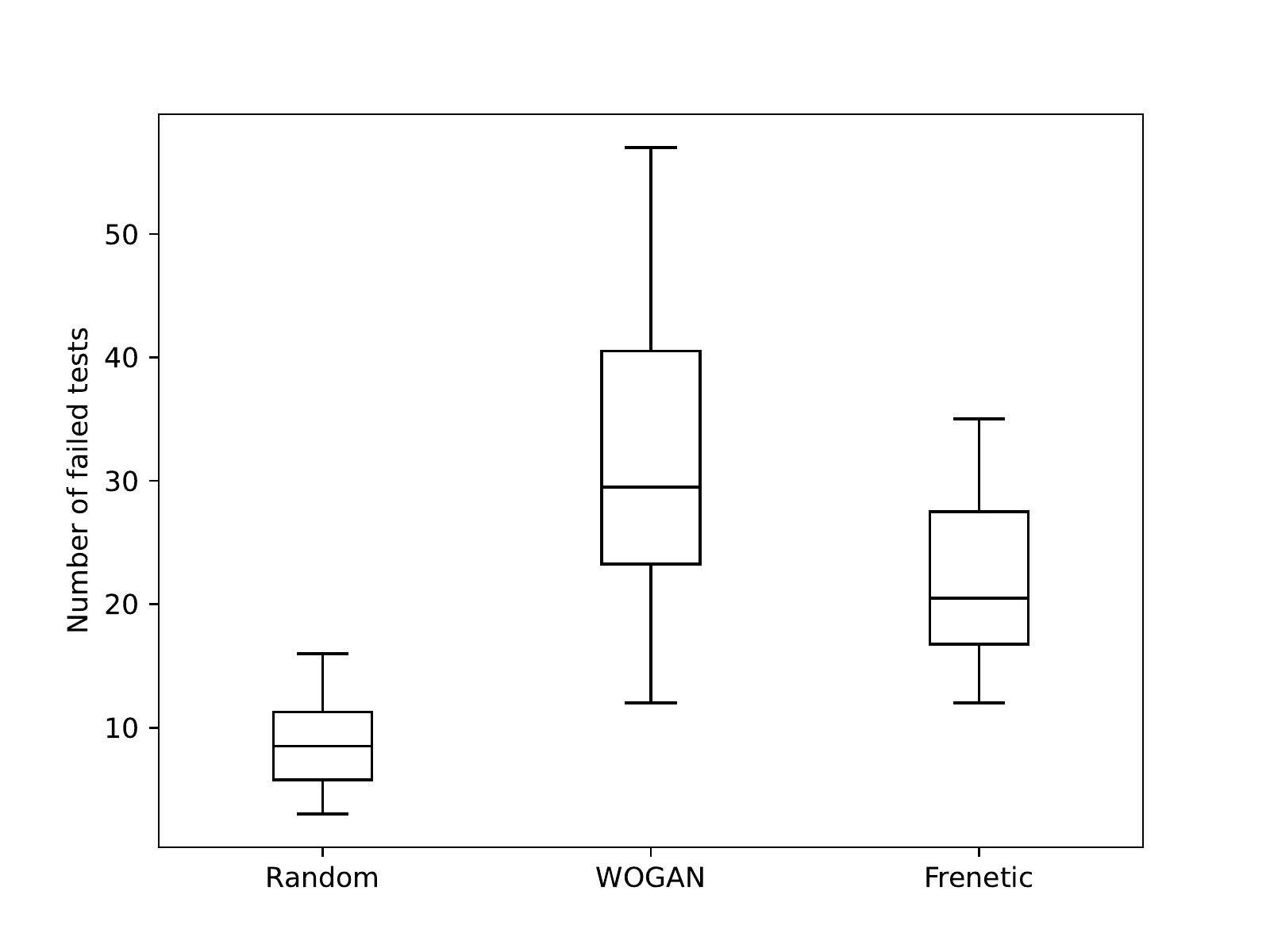}
  \caption{\textmd{Box plots for number of failing tests over $20$ experiments.}}\label{fig:failed}
\end{figure}

\begin{table}
  \caption{\textmd{Table for main statistics of the experiments.}}\label{tbl}
  \begin{tabular}{c|c|c|c}
                                        & \textbf{Random}  & \textbf{WOGAN} & \textbf{Frenetic} \\ \hline
    \textbf{mean executed tests}        &  $337.7$         & $253.8$        & $201.25$          \\
    \textbf{SD executed tests}          &  $2.28$          & $14.54$        & $11.56$           \\
    \textbf{mean failing tests}         &  $8.70$          & $31.30$        & $21.50$           \\
    \textbf{SD failing tests}           &  $3.81$          & $11.61$        & $6.74$            \\
    \textbf{mean fitness final $80 \%$} &  $0.25$          & $0.50$         & $0.65$            \\
    \textbf{SD fitness final $80 \%$}   &  $0.01$          & $0.04$         & $0.03$            \\
    \textbf{mean fitness final $20 \%$} &  $0.26$          & $0.59$         & $0.68$            \\
    \textbf{SD fitness final $20 \%$}   &  $0.03$          & $0.03$         & $0.04$            \\
    \textbf{mean diversity}             & $16.41$          & $11.51$        & $16.85$           \\
    \textbf{SD diversity}               &  $4.66$          & $4.59$         & $2.05$            \\
  \end{tabular}
\end{table}

\noindent
\textbf{Number of Failing Tests.}
See \autoref{fig:failed} for box plots for the number of failing tests in each of the $20$ experiments per algorithm.
See also \autoref{tbl} for key statistics. Evidently both WOGAN and Frenetic beat the random search which can only find
$8.7$ failing tests on average out of an average of $338$ tests executed per experiment. WOGAN is statistically
different from Frenetic (at significance level $0.05$): the Wilcoxon signed-rank test reports a $p$-value of $0.02$
under the null hypothesis that the median of differences is $0$. In \autoref{tbl}, we also include the means of average
fitnesses of the final $80 \%$ and final $20 \%$ of the test suites. Thus WOGAN and Frenetic manage not only to find
failing tests but high-fitness tests in general.

The results indicate that our approach can achieve performance comparable to an
algorithm using more domain-specific knowledge. Thus we have experimentally validated that our WOGAN algorithm fulfills the first quality criterion of
finding many failing tests. \\

\noindent
\textbf{Generation Time.}
We define the generation time of a test being the time elapsed between two executions of a valid test. Average
generation times are $0.5 \, \textup{s}$ (SD $0.2 \, \textup{s}$), $5.1 \, \textup{s}$ (SD $5.2 \, \textup{s})$, and
$3.3 \, \textup{s}$ (SD $11.2 \, \textup{s}$) for Random, WOGAN, and Frenetic respectively. However, generation time is
mostly spent on testing if a candidate test is valid. The remaining time is
negligible for Random and on average $146.7 \, \textup{ms}$ (SD $14.8 \, \textup{ms}$) and $1.4 \, \textup{ms}$ (SD
$1.5 \, \textup{ms}$) for WOGAN and Frenetic respectively. WOGAN uses this time  mainly for model training. All times
reported are measured in real time. Observe that WOGAN manages to execute more tests than Frenetic (see \autoref{tbl})
even though it uses more time for generation. This is explained by noting that Frenetic produces longer roads which
require longer simulation time.\\

\noindent
\textbf{Failing Test Diversity.}
It is an open problem how to measure if the failing tests are ``diverse''. The SBST 2021 CPS testing competition
\cite{DBLP:conf/sbst/PanichellaGZR21} used a certain notion of sparseness to measure this, but here we opt for the
following simpler measure. For each test, we find its interpolation to a road (this is what is actually fed to the
simulator) and evenly reduce it to obtain a sequence of plane points of length $75$ (the minimum length we observed).
These points are rotated in such a way that the initial direction is always directly up. Then we transform the point
sequence to angles between two consecutive points to obtain a vector in $\R^{74}$. For a complete test suite, we define
its diversity to be the median of the pairwise Euclidean distances of these vectors for failing tests. Small diversity
value indicates that the failing tests are not diverse.

\begin{figure}
  \includegraphics[width=\columnwidth]{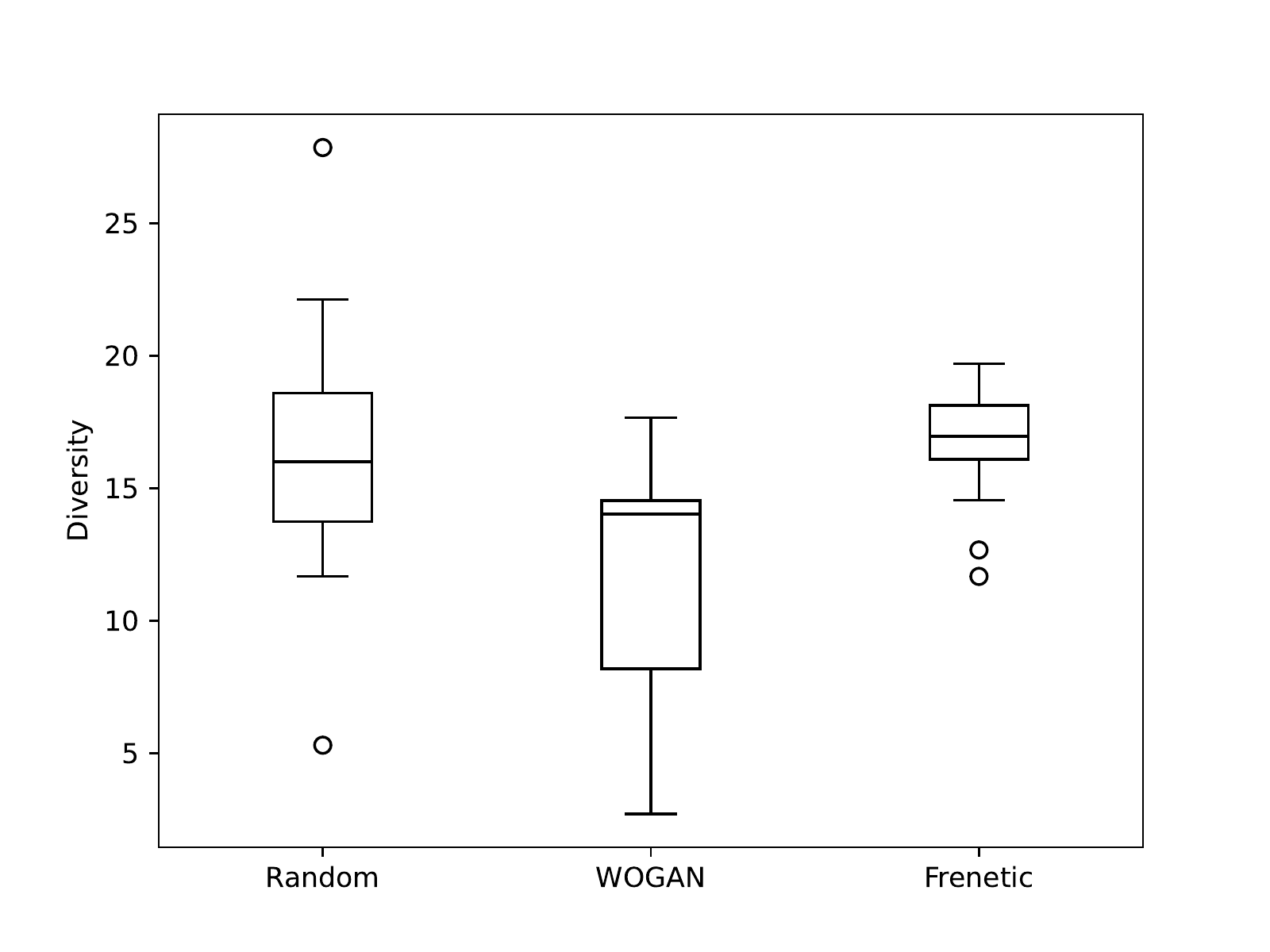}
  \caption{\textmd{Box plots for test suite diversities over $20$ experiments.}}\label{fig:diversity}
\end{figure}

The diversities of the test suites are reported in \autoref{fig:diversity} and additional statistics are found in
\autoref{tbl}. Intuitively the set of failing tests for random search should be diverse. Therefore
\autoref{fig:diversity} indicates that WOGAN and Frenetic both succeed at producing reasonably diverse failing tests
according to the selected diversity measure. The diversity values for WOGAN and Frenetic are statistically different
(at significance level $0.05$): the Wilcoxon signed-rank test reports a $p$-value of $0.0001$ under the null hypothesis
that the median of differences is $0$. We remark that the roads generated by Frenetic are based on a higher number of
plane points, so they are by nature more varied. We conclude that we have shown evidence that WOGAN fulfills the second
quality criterion of being able to generate diverse failing tests. It is true that the data suggests that WOGAN removes
diversity, but accurately assessing the situation would require a deeper study of the tradeoffs between our conflicting
quality criteria.

\section{Conclusions}\label{sec:conclusions}
We have presented a novel online black-box test generation algorithm based on ideas from generative adversarial
networks. The algorithm uses no explicit domain knowledge except in the choice of problem feature representation.
Additionally, the algorithm learns a generator which can be thought of as a model for high-fitness tests for the SUT.
Further examination of this model could prove useful in studying the SUT in more detail. 

We have shown experimentally that our WOGAN algorithm can achieve a  performance comparable to previous competitive
algorithms. Not only is WOGAN able to find faults of the SUT, but it can produce a varied set of failing tests.
However, we remark that our evaluation is based on a single experiment and that the merits of the WOGAN algorithm
should be examined independently in the context of the SBST 2022 CPS testing competition.\footnote{\textbf{Note added in proof:} The competition report is available at \cite{SBST-toolcomp22}. See also \cite{sbst_short} for a brief note on WOGAN's performance in the competition.}

In the future, we aim to write a more complete subsequent work providing full details, additional experiments, and
actual code.

\begin{acks}
This research work has received funding from the ECSEL Joint Undertaking (JU) under grant agreement
No 101007350. The JU receives support from the European Union’s Horizon 2020 research and innovation
programme and Sweden, Austria, Czech Republic, Finland, France, Italy, Spain.
\end{acks}

\bibliographystyle{ACM-Reference-Format}
\bibliography{sbst}

\end{document}